\documentclass[10pt,twocolumn,letterpaper]{article}

\usepackage{wacv}              % To produce the CAMERA-READY version
\usepackage{multirow}
\usepackage[table]{xcolor}
\usepackage{colortbl}
\usepackage[accsupp]{axessibility}
\definecolor{wacvblue}{rgb}{0.21,0.49,0.74}
\usepackage[pagebackref,breaklinks,colorlinks,allcolors=wacvblue]{hyperref}

\def\wacvPaperID{688} % *** Enter the WACV Paper ID here
\def\confName{WACV}
\def\confYear{2026}

\title{Intra-Class Probabilistic Embeddings for Uncertainty Estimation in Vision-Language Models}

\author{
Zhenxiang Lin\and
Maryam Haghighat\and
Will Browne\and
Dimity Miller\and
{ Queensland University of Technology, Brisbane, Australia }\\
{\tt\small \{z25.lin, maryam.haghighat, will.browne, d24.miller\}@qut.edu.au}
}

\begin{document}

% \makeatletter
% \let\@oldmaketitle\@maketitle% Store \@maketitle
% \renewcommand{\@maketitle}{
%    \@oldmaketitle% Update \@maketitle to insert...
%  \begin{center}
%       \includegraphics[width=1\linewidth]{src/teaser.png}
%  \end{center}
%   \refstepcounter{figure}\normalfont Figure~\thefigure. 
%   Top-1 prediction is not enough for a safety-critical robotic perception. Even when the top prediction is correct, inconsistent or semantically unrelated top-k predictions may reveal the unreliablilty of understanding. We highlight the importance of assessing top-k semantic coherence as a complementary signal for reliable visual perception.
%   \label{fig:teaser}
%   \newline
%   }
% \makeatother

\maketitle
\begingroup
\renewcommand\thefootnote{}
\footnotetext{The authors acknowledge continued support from the Queensland University of Technology (QUT) through the Centre for Robotics.}
\addtocounter{footnote}{-1}
\endgroup
\begin{abstract}

Vision-language models (VLMs), such as CLIP, have gained popularity for their strong open vocabulary classification performance, but they are prone to assigning high confidence scores to misclassifications, limiting their reliability in safety-critical applications. We introduce a training-free, post-hoc uncertainty estimation method for contrastive VLMs that can be used to detect erroneous predictions. The key to our approach is to measure visual feature consistency within a class, using feature projection combined with multivariate Gaussians to create class-specific probabilistic embeddings. 
% \color{blue}
% At inference, test embeddings are scored using the log-probability under each class distribution, with a softmax-normalized density used as a distributional confidence score. This score is then combined with the max softmax confidence score derived from cosine similarity to generate a new uncertainty score. 
% \color{black}
Our method is VLM-agnostic, requires no fine-tuning, demonstrates robustness to distribution shift, and works effectively with as few as 10 training images per class. Extensive experiments on ImageNet, Flowers102, Food101, EuroSAT and DTD show state-of-the-art error detection performance, significantly outperforming both deterministic and probabilistic VLM baselines. 
% Our code will be made publicly available upon acceptance.
Code is available at \href{https://github.com/zhenxianglin/ICPE}{https://github.com/zhenxianglin/ICPE}.

\end{abstract}
    
\section{Introduction}

Contrastive vision-Language Models (VLMs), such as CLIP \cite{clip} and SigLIP \cite{siglp}, have achieved remarkable performance across a wide range of visual recognition tasks by learning from internet-scale datasets to align images and textual concepts in a shared embedding space. This has led to rapid adoption for open-vocabulary perception in safety-critical applications, such as medical diagnostics \cite{tiu2022expert, zhang2024mediclip}, robot navigation \cite{gadre2023cows, huang2022visual} and robot action planning \cite{ji2025robobrain, dang2023clip}.
% \TODO{List some applications, e.g. medical diagnostics, perception for self-driving cars?, robot navigation, etc. and add citations for them}.
However, with real-world deployment, model errors have the potential for harmful consequences, and so models must also output a measure of confidence in their predictions.

Contrastive VLMs predict class labels by computing the cosine similarity between an image embedding and the textual prompt for each class \cite{clip, align, siglp, coca, imagebind, languagebind}. 
% While this allows for adaptive querying for different classes and is computationally efficient, it is easy and efficient, this mechanism lacks calibrated uncertainty. 
This cosine similarity, or its softmax transformation, is often used to represent prediction confidence \cite{farina2024zero, trustvlm}. Recent work has shown that this is not a reliable measure of prediction confidence, as VLMs can produce high cosine similarity scores even when an input is misclassified as an incorrect semantic class \cite{tu2023closer, miller2024open, wang2023clipn, levine2023enabling}. In this paper, we address the task of error detection for contrastive VLMs, introducing a technique to produce a new confidence score that can be used to identify errors in classification.

% Such overconfident mistakes are especially problematic when users rely on model confidence to detect errors or abstain from uncertain predictions. This motivates the requirement for reliable uncertainty estimation techniques.
% Error detection has been extensively studied in traditional vision models \cite{hendrycks2016baseline, 10094135, du2022siren}, typically by producing a measure of \emph{uncertainty} for each prediction that can then be used to flag potential misclassifications.
Importantly, prior work has shown that contrastive VLMs suffer from a modality gap where image and text embeddings occupy disjoint regions of the feature space~\cite{liang2022mind}. This suggests that inter-modal similarity measures may not reliably capture prediction uncertainty for identifying errors, as distances between image and text features may be dominated by the modality gap itself rather than by true semantic uncertainty. Building on this observation, \citet{trustvlm} demonstrated that using deterministic visual prototypes for image–image similarity offers stronger error detection than image–text cosine similarity, highlighting the benefit of extracting uncertainty within the visual modality.

% Error detection has been extensively studied in traditional vision models \cite{hendrycks2016baseline, 10094135, du2022siren}, typically by producing a measure of \emph{uncertainty} for each prediction that can then be used to flag potential misclassifications. 
% However, these techniques were originally developed for architectures that rely on linear layers to project features into a fixed set of class scores~\cite{simonyan2014vgg, he2016resnet, dosovitskiy202vit} and do not always clearly translate to the contrastive classification mechanism used by VLMs.
% Classical calibration and uncertainty estimation methods, such as temperature scaling\cite{pmlr-v70-guo17a}, Monte Carlo Dropout\cite{gal2016dropout}, and deep ensembles\cite{lakshminarayanan2017simple}, have shown considerable success. Although these approaches can be adapted to vision-language models (VLMs), their effectiveness remains limited, as they were not designed for multimodal models.
Other recent works have introduced  uncertainty estimation methods for VLMs by learning probabilistic embeddings that model inter-modal uncertainty (i.e. image-text alignment)~\cite{chun2021pcme, chun2024pcmepp, upadhyay2023probvlm, ji2023map, chun2025prolip, baumann2024bayesvlm}. While this improves alignment in the embedding space and improves classification accuracy, it does not necessarily capture uncertainty for error detection (see Sec.~\ref{sec:sotaresults}). Additionally, these techniques often rely on large-scale pretraining or fine-tuning of VLMs, making them either computationally expensive \cite{chun2025prolip, ji2023map} or restricting their generalization beyond the fine-tuning dataset \cite{upadhyay2023probvlm, chun2021pcme, chun2024pcmepp}.

% These approaches can broadly be categorized into deterministic and probabilistic approaches. Deterministic approaches \cite{farina2024zero, trustvlm} typically estimate uncertainty by calibrating prediction scores based on test-time augmentations or feature similarity. P
% probabilistic approaches \cite{chun2021pcme, chun2024pcmepp, upadhyay2023probvlm, ji2023map, chun2025prolip, baumann2024bayesvlm} to estimate prediction uncertainty, modeling learnt distributions over image and text inputs in the embedding space.

% Moreover, most existing methods emphasize modeling inter-modal uncertainty (i.e. image–text alignment) rather than intra-class variability within the visual domain, limiting their effectiveness for tasks like error detection.

\begin{figure*}[h] 
  \centering
  \includegraphics[width=0.991\textwidth]{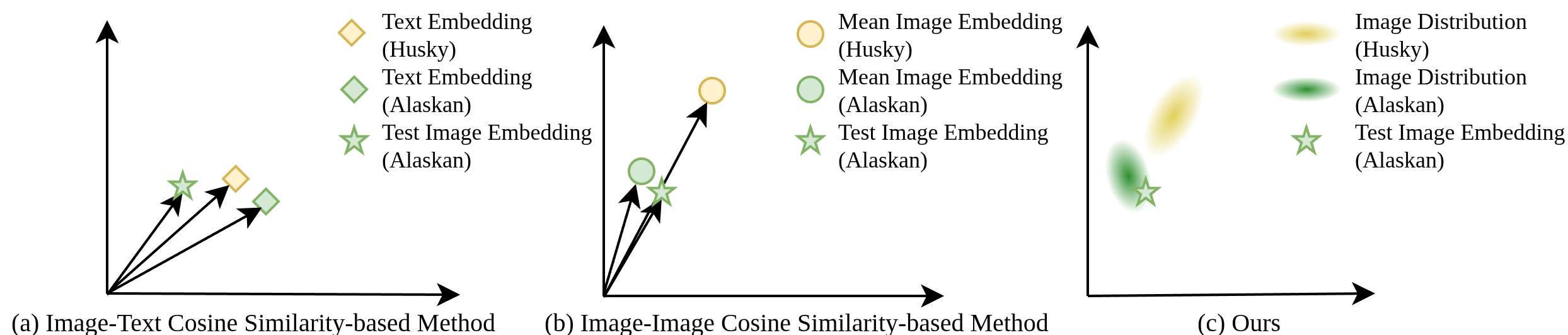}
  \caption{Illustration of different uncertainty estimation paradigms for vision-language models (VLMs). (a) The standard image–text cosine similarity approach assigns high confidence to the Alaskan image (green star) due to its high similarity with the "Husky" text embedding, despite being a misclassification. (b) Image–image similarity methods estimate uncertainty based on proximity to the class mean (yellow circle), but ignore the feature distribution, leading to unreliable scores. (c) Our method models intra-class distributions using image features and assigns higher uncertainty when a test sample deviates from the class distribution, enabling more accurate uncertainty estimation.}
  \label{fig:teaser} 
\end{figure*}

Motivated by this, we propose a technique for training-free, post-hoc uncertainty estimation for error detection in VLMs. Rather than modeling inter-modal uncertainty, we introduce probabilistic embeddings that capture intra-class feature distributions in the visual embedding space (see Fig.~\ref{fig:teaser}). Leveraging training data, we fit a multivariate Gaussian for each class to represent the range of typical visual features. At inference, we retrieve the class distributions and compute the log-probability of the test image under each Guassian, producing an uncertainty score that reflects how well the visual features agree with known intra-class structure.

% The key insight behind our technique is to introduce probabilistic embeddings that model \emph{intra-class feature distributions} in the visual embedding space (see Fig.~\ref{fig:teaser}). Leveraging data from training datasets, for each class we build a multivariate Gaussian distributions representing the range of typical feature embeddings. At inference, we then retrieve the relevant probabilistic embedding for all query classes and compute the log-probability density of the test image under each Gaussian distribution. This produces an uncertainty score that captures the degree of agreement between the test image and the predicted class distribution.

We make the following claims:
\begin{itemize}
  \item Compared to deterministic \cite{farina2024zero, trustvlm} and inter-modal uncertainty estimation techniques \cite{chun2024pcmepp, chun2025prolip, baumann2024bayesvlm, trustvlm}, our intra-class probabilistic embeddings achieve state-of-the-art performance for VLM error detection (see Sec.~\ref{sec:sotaresults}). Our approach is post-hoc and training-free, and can obtain state-of-the-art performance with multiple VLM backbones with as few as 10 images per class (see Sec.~\ref{sec:data-dependency}).
  \item We identify that PCA-based feature projection is crucial for modeling each class distribution effectively (see Sec.~\ref{sec:pca}), positing that this mitigates ill-conditioned covariance matrices in the VLM feature space. 
  % \item Our approach exhibits robustness to label shift, and can continue to achieve state-of-the-art performance even when the probabilistic embedding labels have some discrepancy with the test query labels (see Sec.~\ref{sec:label-shift}).
  \item Our approach exhibits robustness to both label shift (semantic mismatches between query labels and dictionary labels) and feature shift (distributional differences between training and test images) (see Sec.~\ref{sec:dist-shift}).
 
\end{itemize}

\section{Related Works}
% \TODO{Related Works}
\subsection{Uncertainty Estimation for Vision Models}
Detecting misclassifications in image classifiers is a key challenge in computer vision. A common method is the maximum softmax probability (MSP), which use the model's most confident prediction as a proxy for correctness \cite{hendrycks2016msp}. However, deep networks are prone to overconfidence \cite{wang2023clipn, tu2023closer}, often assigning high confidence to incorrect predictions.

To address this, several works \cite{pmlr-v70-guo17a, liang2017enhancing} apply temperature scaling to calibrate confidence. Differently, Bayesian approximations-based methods \cite{graves2011practical, blundell2015weight, louizos2017bayesian, rezende2015variational, gal2016dropout} treat model parameters as probability distributions and introduce uncertainty directly into the model weights. Ensemble-based methods \cite{lakshminarayanan2017simple, achrack2020multi, valdenegro2019deep, wen2020batchensemble} train multiple independent models and combine the variance from their predictions to quantify uncertainty. Nonetheless, these methods still require high computational costs due to the training and testing of multiple models. Single network deterministic
methods \cite{malinin2018predictive, sensoy2018evidential, raghu2019direct, van2020uncertainty} incorporate uncertainty estimation in a single deterministic forward pass and predict the distribution parameters of the output.

These methods have achieved strong performance in  confidence calibration and misclassification detection across a variety of image classification tasks. However, they have been primarily developed for conventional vision-only architectures, such as CNNs and Transformers, where prediction is made solely based on visual inputs. For VLMs, like CLIP \cite{clip} and SigLIP \cite{siglp}, which perform classification by computing similarities between image and text embeddings, these techniques are not directly applicable. 

\begin{figure*}[h] % here/top/bottom/page
  \centering
  \includegraphics[width=1.0\textwidth]{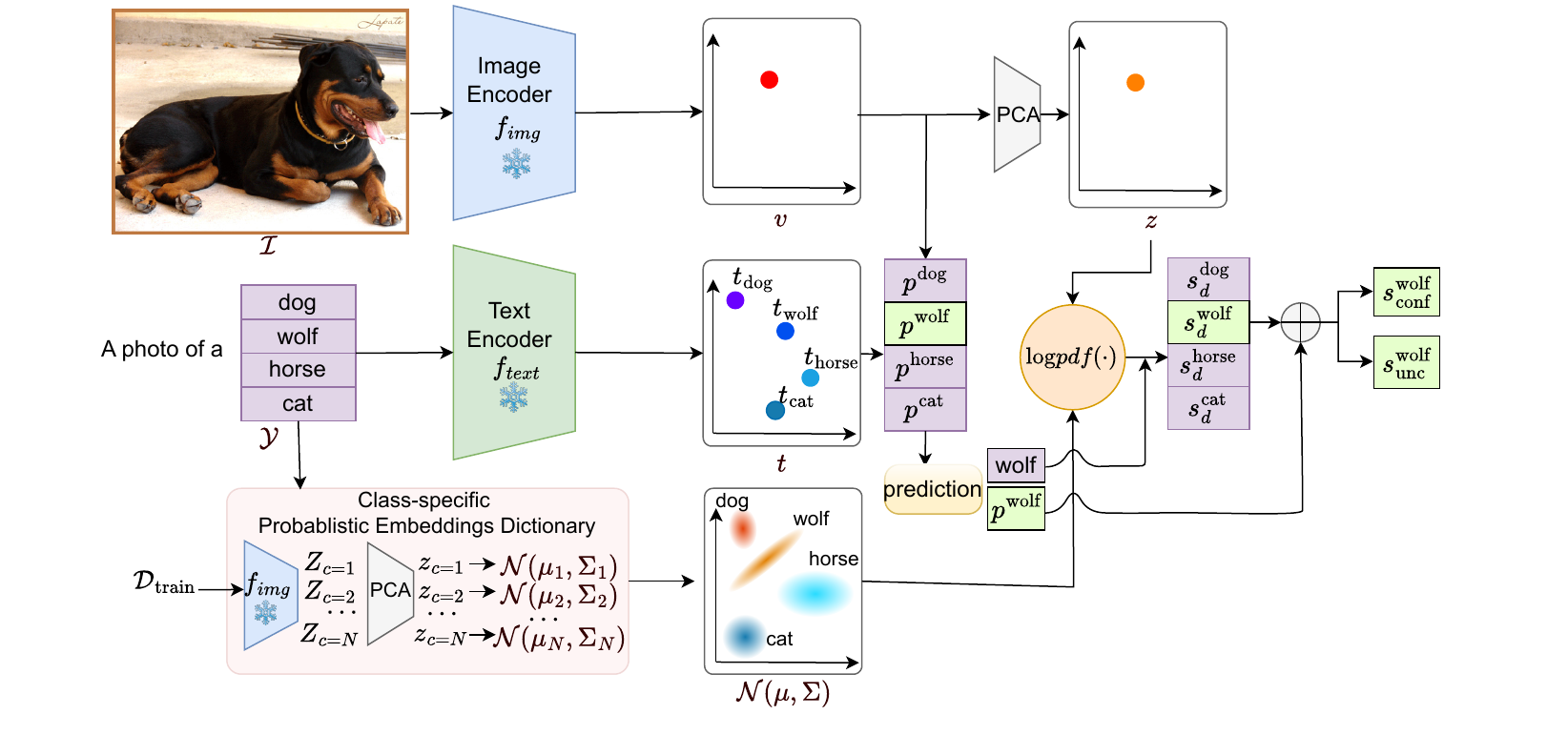}
  \caption{Overview of the proposed retrieval-augmented uncertainty estimation pipeline, which utilises a dictionary of intra-class probabilistic distributions to estimate uncertainty.
  % . A test image is first encoded via a frozen image encoder and compared to class text embeddings using cosine similarity to obtain the prediction and its corresponding max softmax value $p$. Meanwhile, a set of training images is encoded to generate the class-specific probabilistic embeddings. The test image’s embedding is also projected by PCA and scored under the predicted class’s distribution using log probability density. 
  % \color{blue}Then, the distributional confidence score $s_d$ is obtained by applying softmax over class-wise log probability densities. Finally, the uncertainty and confidence scores are computed by summing the two sources of confidence. \color{black}
  % the  \todo{need to adjust to show summing softmax score of image-text match.}
  % \TODO{retriever changed to dictionary, class embeddings should have different levels of variance. Show also the pre-inference setup of probabilistic embedding dictionary as described in 3.3. Tidy this figure up -- rather than (Frozen) you can put a frozen symbol. Use notation consistent with the methodology section.The cosine similarity + softmax part looks clunky and not neat., as does the Pred: wolf bit}
  }
  \label{fig:pipeline} 
\end{figure*}

\subsection{Uncertainty Estimation for VLMs}
VLMs \cite{clip, align, siglp, coca, languagebind, imagebind} have demonstrated impressive performance on a variety of visual recognition tasks through their ability to align images and textual concepts in a shared embedding space. In classification, VLMs typically assign labels by computing the cosine similarity between the image embedding and each class's textual prompt, and then applying softmax over the similarity scores. However, unlike traditional vision-only models, which rely on an explicit classifier, this similarity-based prediction mechanism raises fundamental challenges for estimating uncertainty.

To address this, Zero \cite{farina2024zero} uses the test-time adaptation strategy to calibrate confidence. However, it relies on multiple forwards at inference, causing a low inference speed. PCME \cite{chun2021pcme}, PCME++ \cite{chun2024pcmepp} and ProbVLM \cite{upadhyay2023probvlm} use probabilistic adapters to model distributions over text and image embeddings, but they require finetuning on a dataset for a specific task. MAP \cite{ji2023map} and ProLIP \cite{chun2025prolip} proposed pre-training models for probabilistic representation that can benefit many downstream tasks. However, these methods need to cost a large number of computation sources and rely on an Internet-scale dataset. BayesVLM \cite{baumann2024bayesvlm} leverages a posterior Laplace approximation to quantify uncertainties over cosine similarities, but its dependence on Hessian-based estimation requires access to the underlying loss function and model gradients, which may not always be accessible for all VLMs. TrustVLM \cite{trustvlm}, which is a concurrent work to ours, takes visual feature consistency into consideration and explore the error detection task, but they do not model a distribution and instead have a single prototype per class.
% estimates confidence by image-image cosine similarity, while they ignore the distribution of each class. 
Our method leverages class-wise distribution modeling over frozen VLM features to assess the distributional typicality of predictions, achieving effective uncertainty estimation without retraining or architectural modification.

% \subsection{Retrieval-Augmented Modeling}

% \vspace{-1ex}
\section{Methodology}
\subsection{Preliminaries of VLM Classification}
\label{sec:vlm-cls}
Contrastive VLMs, such as CLIP~\cite{clip}, encode both images and natural language into a shared feature space using modality-specific encoders. VLMs can be used for zero-shot classification by comparing the similarity between the encoded image and a set of class-descriptive text prompts.

Let $\mathcal{I}$ denote an input image and $\mathcal{Y} = \{y_1, \ldots, y_K\}$ the set of $K$ class labels. Each image $\mathcal{I}$ is encoded into a visual feature vector $v = f_{\text{img}}(\mathcal{I}) \in \mathbb{R}^d$ using a visual encoder $f_{\text{img}}$. Each class label $y_k$ is formatted into a natural language prompt (e.g., ``a photo of a \texttt{<label>}''), and then encoded into a text feature $t_k = f_{\text{text}}(y_k)$ using a text encoder $f_{\text{text}}$.

Classification is performed by computing the cosine similarity between the image embedding $v$ and each text embedding $t_k$:
\[
s_k = \frac{v \cdot t_k}{\|v\| \, \|t_k\|}, \quad \text{for } k = 1, \ldots, K.
\]
The predicted class $\hat{y}$ corresponds to the label with the highest similarity:
\[
\hat{y} = \arg\max_{k} \, s_k.
\]

The similarity scores $\{s_k\}_{k=1}^K$ can also be interpreted as a measure of confidence or uncertainty. Typically, these scores are normalized using a softmax function:
\[
p_k = \frac{\exp(s_k)}{\sum_{j=1}^K \exp(s_j)},
\]
and the maximum probability or the entropy of the distribution can also serve as a confidence or uncertainty estimate. However, prior work has shown that VLM's can assign high cosine similarity even when an input is misclassified~\cite{trustvlm, farina2024zero}, making these cosine-based uncertainty measures suboptimal for error detection.

% \TODO{Methodology}

% \subsection{Limitation of Cosine-based Uncertainty}
% Vision-language models \cite{clip, align, siglp, coca, imagebind, languagebind} produce classification predictions by computing the cosine similarity between an image embedding and a set of text embeddings representing class labels. These similarity scores, when passed through a softmax function, are often interpreted as uncertainty estimates. However, in practice, these scores are poorly calibrated and frequently lead to overconfident or underconfident predictions \cite{tu2023closer, trustvlm}. In particular, they tend to reflect how well an image aligns with a single text prompt, but fail to reflect the extent to which an image visually matches an individual class.

% This limitation is especially problematic in high-stakes or uncertainty-sensitive applications. A prediction may exhibit high cosine similarity to the text embedding of a certain class, but still be incorrect, especially when the image is visually misleading. The softmax confidence in such cases does not account for whether the image "looks like" other images from that class in a distributional sense. This reveals a fundamental shortcoming of cosine-based confidence estimation: it measures pairwise alignment, but not class consistency. We argue that more reliable uncertainty estimation should assess how typical a prediction is relative to the overall feature distribution of the predicted class, rather than relying solely on image-text similarity.

\subsection{Method Pipeline}
% To address the limitations of cosine-based uncertainty estimation, we propose an approach that estimates uncertainty based on the feature consistency of the test image with existing images from the predicted class. 
An overview of our approach is shown in Figure~\ref{fig:pipeline}. Utilizing existing training datasets, our approach creates a dictionary of class-specific probabilistic distributions of feature embeddings for all labeled classes.
This captures class-level structure in the VLM feature space, which is absent in the standard cosine similarity scores. 

% Given a test image, we compute the log-probability density of its feature embedding (after projection) under the distribution of the predicted class. After softmax, the resulting values are used as the prediction uncertainty for each category. Atypical images, which deviate from the expected distribution of the predicted class, can receive high uncertainty even if their cosine similarity is high.

At inference, given a test image, we first extract its deterministic visual feature and obtain its predicted class label (following the standard approach introduced in Sec.~\ref{sec:vlm-cls}). We then project the feature to a low dimension space and compute the log-probabilities of the projected embedding under the Gaussian distribution of all queried classes in our dictionary. The softmax-normalized log-probability of the predicted class is combined with the image-text softmax score and used to represent prediction uncertainty, where predictions with high uncertainty are flagged as likely errors. Notably, our method does not alter the model's original prediction and introduces no training. It operates entirely on frozen features, providing a post-hoc uncertainty estimate that is agnostic to VLM backbone.

\subsection{A Dictionary of Class-specific Probabilistic Embeddings}
We assume access to a dataset $\mathcal{D}_{\text{train}} = \{(\mathcal{I}_i, y_i)\}_{i=1}^N$ consisting of $N$ images $\mathcal{I}_i$ paired with class labels $y_i \in \mathcal{Y}_{\text{train}}$. Importantly, the visual features from $\mathcal{D}_{\text{train}}$ should be independent and identically distributed (i.i.d.) with the visual features of test images. In practice, this can be achieved by using the training subset for any given test dataset. While our method assumes i.i.d. visual features, it does not require $\mathcal{Y}_{\text{train}}$ to identically match the test label set. In Section~\ref{sec:dist-shift}, we show how our approach can handle moderate class label shift and feature shift.
% ---i.e., where the test-time classes $\mathcal{Y}_{\text{test}}$ differ partially from $\mathcal{Y}_{\text{train}}$.

% Prior to inference, we construct a dictionary of class-specific probabilistic embeddings that represent the full visual distributional structure of a class in the VLM feature space. This allows our method to evaluate whether a test image resembles other images of that class in a statistically meaningful way.

To construct our dictionary of class-specific probabilistic distributions, we iterate over each class label $c \in \mathcal{Y}_{\text{train}}$. For each class $c$, we retrieve all training images from $\mathcal{D}_{\text{train}}$ that are labeled with $c$, and pass them through the frozen visual encoder of the VLM to obtain a set of high-dimensional feature vectors. We denote this set of visual embeddings as $Z_c = \{\mathbf{v_i} \,|\, y_i = c\}$. As motivated in Section~\ref{sec:pca}, we then apply principal component analysis (PCA) to reduce the dimensionality of all embeddings before fitting class-conditional distributions.

We treat the PCA-transformed embeddings in $Z_c$ as i.i.d.\ samples drawn from a multivariate distribution over the reduced feature space. Each class $c$ becomes a key in our dictionary, associated with a multivariate Gaussian probabilistic embedding, parameterized by the mean $\mu_c$ and covariance matrix $\Sigma_c$:
\[
\mu_c = \frac{1}{|Z_c|} \sum_{\mathbf{v} \in Z_c} \mathbf{v}, \quad
\Sigma_c = \frac{1}{|Z_c| - 1} \sum_{v \in Z_c} (\mathbf{v} - \mu_c)(\mathbf{v} - \mu_c)^\top.
\]

% To build our dictionary, We adopt a retrieval-augmented approach, where each class distribution is constructed from a collection of images that share the same ground-truth label. Specifically, given a set of candidate labels ${c_1, c_2,…,c_N}$, we retrieve for all corresponding images for each class $c$. These images are passed through the frozen VLM image encoder to obtain a set of high-dimensional embeddings $Z_c$:
% \begin{equation}
% \begin{aligned}
%     Z_c = f_I(I_c),
% \end{aligned}
% \label{eq:visual-embedding}
% \end{equation}
% where $f_I$ is image encoder and $I_c$ is all corresponding images for class $c$.

% This label-supervised retrieval strategy avoids the noise and ambiguity associated with similarity-based retrieval. It ensures that the modeled feature distribution for each class is fully aligned with the semantics of the classification task.

% To capture the distribution of each class, we treat the retrieved feature set $Z_c$ as samples from a multivariate distribution. We model this distribution using a class-conditional multivariate Gaussian, parameterized by its empirical mean $\mu_c$ and covariance matrix $\Sigma_c$:
% \begin{equation}
% \begin{aligned}
%     \mu_c &= \frac{1}{|Z_c|}\sum_{z_i \in Z_c}z_i, \\
%     \Sigma_c &=\sum_{z_i \in Z_c}(z_i-\mu_c)(z_i-\mu_c)^T.
% \end{aligned}
% \label{eq:gaussian}
% \end{equation}

\subsection{Feature Projection for Mitigating Ill-conditioned Covariance Matrices}
\label{sec:method-pca}
% \TODO{Check for notation consistency with the rest of the paper}
Although VLMs' image embeddings are powerful for semantic alignment, their distribution has been shown to be anisotropic and highly correlated \cite{levi2024double,tyshchuk2023isotropy}. Specifically, the covariance matrix $\boldsymbol{\Sigma} = \frac{1}{N} \sum_{i=1}^N (\mathbf{v_i} - \boldsymbol{\mu})(\mathbf{v_i} - \boldsymbol{\mu})^\top$ of the feature embeddings  $\mathbf{v_i}$ contains significant off-diagonal entries, indicating strong linear correlations among feature dimensions \cite{levi2024double}. In addition, the eigenvalue spectrum of $\boldsymbol{\Sigma}$ is  highly non-uniform, suggesting extreme variance disparities across directions \cite{levi2024double}.

 Such a covariance structure is ill-conditioned, characterised by  rank deficiency and a high condition number, i.e. $\kappa(\boldsymbol{\Sigma}) = \frac{\lambda_1}{\lambda_d}$ is very large. This makes class-conditional  modeling unstable and often numerically infeasible, especially when the number of class samples is limited relative to embedding dimension. 

To mitigate this, we apply PCA to obtain an orthogonal basis $ \mathbf{U} \in \mathbb{R}^{d \times d} $ such that $\boldsymbol{\Sigma} = \mathbf{U} \boldsymbol{\Lambda} \mathbf{U}^\top$. We  retain only the top-$k$ principal components, and project embeddings to 
$\mathbf{z}_i = \mathbf{U_k}^\top (\mathbf{v_i} - \boldsymbol{\mu})$, where $ \mathbf{U} \in \mathbb{R}^{d \times k} $, and $k\ll d$. The resulting covariance matrix in this projected space is approximately diagonal and significantly better conditioned with a reduced condition number $\kappa(\boldsymbol{\Sigma}) = \frac{\lambda_1}{\lambda_k} \ll \kappa(\boldsymbol{\Sigma})$. The distribution of the projected embeddings can then be approximated as $\mathbf{z}_i \sim \mathcal{N}(\mathbf{0}, \boldsymbol{\Lambda}_k)$  enabling multivariate gaussian modeling. 
% \todo{Explain that we can either find the full covariance matrix, or in the presence of less data, find the diagonal covariance matrix}.
Depending on the amount of data available for Gaussian parameter estimation, our method can either adopt a diagonal covariance approximation or a full covariance approximation \cite{murphy2012ML}. We report results for both variations and explore the trade-off between data availability and the covariance approximation in Sec.~\ref{sec:data-dependency}.

\subsection{Uncertainty Estimation during Inference}
\label{sec:scoring}
With class-wise Gaussian distributions modeled in a stabilized feature space, we introduce an uncertainty scoring method that reflects how well a test image aligns with the feature distribution of its predicted class. 
% Unlike standard softmax scores derived from cosine similarity, our approach leverages intra-class visual consistency to estimate the reliability of a model’s prediction.
Given a test image $\mathcal{I}$, we obtain its visual embedding $\mathbf{v} = f_{\text{img}}(\mathcal{I})$ using the frozen VLM image encoder. The predicted class $\hat{c}$ is then selected by computing cosine similarity between $\mathbf{v}$ and the class text embeddings $\{t_k\}_{k=1}^K$:
\[
\hat{c} = \arg\max_k \langle \mathbf{v}, t_k \rangle.
\]

% The softmax over these similarities provides the model's confidence in each class, maintaining full compatibility with standard VLM inference.

To compute uncertainty, we first project $\mathbf{v}$ into the PCA-transformed space:
\begin{equation}
\mathbf{z} = \mathbf{U_k}^\top (\mathbf{v} - \boldsymbol{\mu}).
\label{eq:test-proj}
\end{equation}

Using the text label for our predicted class $\hat{c}$, we then extract the relevant probabilistic embedding from our dictionary and compute the log-likelihood of $\mathbf{z}$ under the Gaussian distribution of the predicted class:
\begin{equation}
s_{\hat{c}} = \log pdf(\mathbf{z} \mid \mu_{\hat{c}}, \Sigma_{\hat{c}}),
\label{eq:pdf}
\end{equation}
where \( pdf(\cdot) \) is the multivariate Gaussian probability density function.
This score reflects how well the test image aligns with the predicted class’s feature distribution: higher values indicate typicality, while lower values suggest the prediction may be overconfident or erroneous.

To enable uncertainty comparisons across examples, we normalize the log-likelihood scores across all classes using a softmax function to find an intra-class uncertainty score $s_d$.
% \begin{equation}
% s_d = \text{softmax}(s).
% % \quad s_{\text{unc}} = 1 - s_{\text{conf}}.
% \label{eq:unc}
% \end{equation}
% \todo{Add in text here that explains that we add the softmax between image-text with this new score.}
To create our final uncertainty score, we combine this with the VLM inter-modal similarity
\begin{equation}
s_{\text{unc}} = 1 - \frac{p_{\max} + s_d}{2}.
\label{eq:final-score}
\end{equation}
where $p_{\max}$ denotes the maximum softmax probability obtained from image–text similarities. This combination allows our uncertainty score to represent both inter-modal similarity (which VLMs are extensively trained to capture) as well as the intra-class feature consistency.

The uncertainty score $s_{\text{unc}}$ can be used for error detection by applying a simple thresholding strategy. If $s_{\textbf{unc}} > \tau$, where $\tau$ is a user-defined threshold, the model rejects the prediction as a likely misclassification. Our evaluation metrics test over a range of $\tau$, and in Sec.~\ref{sec:sensitive}, we explore the sensitivity of our method to this threshold.

\subsection{Scalability}
An important practical aspect of the proposed method is its scalability to large label spaces. While our dictionary construction only needs to occur once before deployment, the computational cost of this scales linearly with the number of classes and number of image samples per class. The most costly operation is the feature extraction for each image, which requires a forward pass through the VLM visual encoder. This cost can be alleviated by parallelization and GPU-accelerated inference, and in Sec.~\ref{sec:data-dependency} we show that our method can achieve state-of-the-art performance with at least ten labeled images per class. In combination, this enables our method to feasibly scale to large-scale classification scenarios if such compute is available. We note, however, that our approach may be intractable for applications where the dictionary must be computed on a compute-limited edge devices.

\section{Experimental Results}
\subsection{Experimental Setup}
\begin{table*}[ht]
\centering
\setlength{\tabcolsep}{3pt}
{\fontsize{8}{10}\selectfont 
\caption{Our approach outperforms the uncertainty estimation baselines for the CLIP and SigLIP backbones and the task of error detection across all datasets. *TrustVLM additionally uses DinoV2 features alongside VLM features. Ours-D indicates when our method uses a diagonal covariance approximation for the class-wise Gaussian distributions. Best results are shown in bold. 
% \TODO{also bold for acc column, remove year column, amke sure every column has a bold. Bold best clip method and best siglip method separately}
}
\label{tab:comparison}
\begin{tabular}{lc|ccc|ccc|ccc|ccc|ccc}
\hline
& \multirow{3}{*}{\textbf{Method}}   & \multicolumn{3}{c|}{\textbf{ImageNet}} & \multicolumn{3}{c|}{\textbf{Flowers102}} & \multicolumn{3}{c|}{\textbf{Food101}} & \multicolumn{3}{c|}{\textbf{EuroSAT}} & \multicolumn{3}{c}{\textbf{DTD}} \\
& &  AuROC  & AuPR & FPR95  & AuROC & AuPR & FPR95 &   AuROC & AuPR & FPR95  & AuROC & AuPR & FPR95  & AuROC & AuPR & FPR95  \\
& &  ($\uparrow$)  & ($\uparrow$) & ($\downarrow$)  & ($\uparrow$)  & ($\uparrow$) & ($\downarrow$) & ($\uparrow$)  & ($\uparrow$) & ($\downarrow$)  & ($\uparrow$)  & ($\uparrow$) & ($\downarrow$) & ($\uparrow$)  & ($\uparrow$) & ($\downarrow$) \\
\hline
\multirow{9}{*}{\rotatebox[origin=c]{90}{CLIP ViT-B/32}} & MaxCosine  & 65.79  & 74.58  & 86.40   & 72.71  & 82.61  & 80.85 & 74.60  & 92.42  & 81.65 & 45.72 & 35.14 & 97.84 & 66.16 & 58.61 & 86.56 \\
& MaxSoftmax      & 80.71  & 87.43  & 71.38 & 86.26  & 92.21 & 61.11 & 87.05  & 96.70  & 62.96 & 63.25 & 51.62 & 87.97 & 76.98 & 74.32 & 78.49 \\
& Entropy          & 78.27  & 86.01  & 76.75  & 85.22  & 91.56 & 64.70  & 85.56  & 96.33 & 67.26  & 61.03 & 50.28 & 92.07 & 76.43 & 73.92 & 80.14 \\
& TempScaling~\cite{tu2024empirical} & 80.63 & 87.37 & 71.64   & 86.15 & 92.11 & 61.93  & 87.06 & 96.70 & 62.31 & 63.25 & 51.62 & 87.95 & 77.01 & 74.36 & 77.56 \\
& ProbVLM~\cite{upadhyay2023probvlm}         & 50.18  & 61.79  & 95.31 & 43.05  & 53.74 & 97.20  & 52.87 & 82.99 & 95.16   & 42.72 & 30.23 & 95.53 & 46.57 & 39.52 & 97.26 \\
& PCME++~\cite{chun2024pcmepp}          & 77.31  & 90.68 & 71.82 & 80.05  & 93.89  & 70.11 & 87.60  & 97.84 & 53.44 & 89.31	& \textbf{99.80}	& 42.57 & 82.06	& \textbf{93.15}	& 75.65 \\
& BayesVLM~\cite{baumann2024bayesvlm}     & 79.41   & 86.05  & 72.83 & 85.65  & 91.88 & 91.37  & 86.12  & 96.32 & 64.67 & 61.46	& 41.49	& 90.56 & 73.77 & 58.74	& 77.32 \\
 \rowcolor{gray!25} \cellcolor{white} &  Ours-D & 85.80 & 90.53 & 59.02 & \textbf{96.96} & \textbf{98.43} & 14.79 & \textbf{92.01} & \textbf{997.93} & 37.04 & 96.56 & 92.57 & 24.82 & \textbf{91.56} & 89.44 & \textbf{40.23} \\
 \rowcolor{gray!25} \cellcolor{white} &  Ours & \textbf{87.52} & \textbf{91.61} & \textbf{55.26}  & 97.59 & 98.68 & \textbf{4.86} & 92.58	& 98.08	& \textbf{33.98} & \textbf{96.37}	& 94.14 & \textbf{16.39} & 91.13	& 89.09	& 46.43 \\

\hline
% PCME++*~\cite{chun2024pcmepp}        & 2024 & CLIP ViT-B/32   & 79.22  & 92.18  & 71.57 & 77.61  & 80.17  & 93.99  & 70.02 & 82.05  & 89.55  & 98.47  & 54.78 & 89.05  \\
\multirow{9}{*}{\rotatebox[origin=c]{90}{CLIP ViT-B/16}}& MaxCosine & 70.20   & 87.04  & 82.31 & 79.98   & 94.32  & 64.23 & 77.61  & 96.30 & 75.11 & 59.24 & 48.18 & 87.98 & 69.69 & 64.96 & 85.67 \\
& MaxSoftmax & 84.21   & \textbf{94.10}  & 65.30 & 93.49   & 98.64  & 40.18 & 91.43   & 98.92  & 49.98 & 76.21 & 73.69 & 81.15 & 79.96 & 78.54 & 76.33 \\
& Entropy & 83.46   & 93.87  & 68.64 & 93.09   & 98.55  & 36.77 & 90.79   & 98.84  & 54.79 & 71.97 & 67.99 & 83.73 & 79.15 & 77.54 & 76.86 \\
& TempScaling~\cite{tu2024empirical} & 81.28 & 89.87 & 72.16 & 85.41 & 92.60 & 60.58 & 89.21 & 98.12 & 54.61 & 76.22 & 73.70 & 81.28 & 79.94 & 78.48 & 76.33 \\
& Zero~\cite{farina2024zero}           & 78.87  & 90.84  & 73.95 & 83.73  & 92.30  & 62.98 & 85.13  & 97.84 & 56.62 & 66.73	& 61.81	& 87.23 & 76.64	& 77.68	& 81.79
\\
% Zero~\cite{}         & 2024 & CLIP ViT-B/32   & 78.77  & 88.82  & 66.86  &   &   &   &   &   &   \\
& ProLIP~\cite{chun2025prolip}          & 70.57   & 85.77  & 81.23 & 80.67  & 92.82  & 59.95  & 77.05  & 96.81  & 76.04 & 62.42 & 60.64 & 89.44 & 78.96 & 86.70 & 72.74 \\
& TrustVLM*~\cite{trustvlm}   & 83.39 & 90.43  & 75.67  & 89.18  & \textbf{99.86}  & 34.43   & 89.37  & 98.19 & 56.04 & 73.72	& 91.84 & 73.66 & 77.17 & 88.98 & 74.49 \\
 \rowcolor{gray!25} \cellcolor{white} & Ours-D     & 86.80 & 92.57 & 56.57 & 97.28 & 98.79 & 10.62 & 93.40 & 98.82 & 29.03 & 97.32 & 95.91 & 7.39 & \textbf{93.00} & \textbf{91.55} & \textbf{33.86} \\
 \rowcolor{gray!25} \cellcolor{white} & Ours     & \textbf{88.01} & 93.23 & \textbf{54.10} & \textbf{98.25} & 99.15 & \textbf{2.47} & \textbf{94.02} & \textbf{98.94} & \textbf{27.08} & \textbf{97.64} & \textbf{96.16} & \textbf{3.35} & 92.61 & 91.31 & 36.94 \\

\hline
\multirow{8}{*}{\rotatebox[origin=c]{90}{SigLIP ViT-B/16}}& MaxCosine      & 70.20   & 87.04  & 82.31 & 79.98   & 94.32  & 64.23  & 77.61  & 96.30 & 75.11  & 60.47 & 54.20 & 98.93 & 68.82 & 78.99 & 85.19 
 \\
& MaxSoftmax      & 84.21   & 94.10  & 65.30    & 93.49   & 98.64  & 40.18   & 91.43   & 98.92  & 49.98 & 74.91 & 69.24 & 80.30 & 82.05 & 89.21 & 82.29
   \\
& Entropy         & 83.46   & 93.87  & 68.64   & 93.09   & 98.55  & 36.77   & 90.79   & 98.84  & 54.79 & 74.41 & 69.04 & 82.51 & 80.30 & 88.18 & 71.18
 \\
&TempScaling~\cite{tu2024empirical}   & 84.12  & 94.06  & 66.06    & 93.64  & 98.66  & 38.38   & 91.43  & 98.92 & 49.71  & 74.40 & 68.62 & 80.85 & 81.48 & 88.84 & 72.41 \\
& Zero~\cite{farina2024zero}           & 79.43  & 93.55  & 68.65 &  86.30 & 96.84  & 53.09 & 85.23 & 97.96 & 55.76 & 66.63 & 55.79	& 87.95 & 75.65 & 86.51 & 74.22 \\
& BayesVLM~\cite{baumann2024bayesvlm} & 70.06 & 87.84 & 83.01 & 70.59 & 90.49 & 85.19 & 78.17 & 96.58 & 74.28 & 70.54 & 62.52 & 82.52 & 80.06 & 78.40 & 61.16 \\
& TrustVLM*~\cite{trustvlm} & 81.66 & 90.85 & 81.65 & 93.65 & \textbf{99.93} & 29.82 & 90.08 & 98.50 & 53.15 & 74.97 & 92.16 & 74.20 & 82.84 & 92.78 & 79.90
\\
 \rowcolor{gray!25} \cellcolor{white}& Ours-D & 86.17 & 94.80 & 56.88 & \textbf{99.38} & 99.88 & \textbf{1.40} & \textbf{95.06} & \textbf{99.37} & 28.13 & 96.93 & 96.95 & 12.67 & \textbf{94.08} & \textbf{96.22} & \textbf{24.88} \\
 \rowcolor{gray!25} \cellcolor{white}& Ours & \textbf{88.57} & \textbf{95.85} & \textbf{48.87} & 98.81 & 99.77 & 1.60 & 94.96 & 99.35 & \textbf{24.50} & \textbf{98.10} & \textbf{98.03} & \textbf{2.93} & 89.44 & 94.27 & 68.58 
   \\

\hline
\end{tabular}
}
\end{table*}

\textbf{Metrics:} We treat the error detection task as a binary classification problem (i.e. correct predictions to retain versus error to reject) and adopt three standardized metrics used for this task~\cite{hendrycks2016baseline, 10635759, Park_2023_ICCV, farina2024zero, trustvlm}: Area Under the Receiver Operating Characteristic Curve (AuROC), the False Positive Rate at 95\% True Positive Rate (FPR95) and the Area Under the Precision-Recall Curve (AuPR). We additionally report $\text{top-1}$ accuracy on the multi-class classification task in the Supplementary Material, noting that this is a secondary metric and it is not our goal to improve accuracy.

% Both AuROC and AuPR reflect the performance of an uncertainty score to distinguish between correct and misclassification predictions using binary classification across all thresholds. A value of one indicates perfect distinction between the uncertainty scores for correct and misclassified samples. FPR at 95% TPR measures the false positive rate (i.e., fraction of incorrect predictions incorrectly accepted) when the true positive rate (i.e., fraction of correct predictions correctly accepted) is fixed at 95%
\noindent\textbf{Datasets:}
 We evaluate our method on five standard image classification benchmarks with different domains and scales: ImageNet-1K~\cite{deng2009imagenet} (1000 categories, generic objects and animals), Flowers-102 \cite{nilsback2008flower102} (102 categories, flower species), Food-101 \cite{bossard2014food101} (101 categories, food types), EuroSAT~\cite{helber2019eurosat} (10 categories, satellite imagery and land types) and DTD~\cite{cimpoi14dtd} (47 categories, textures). For each dataset, we use the official training sets to model our probabilistic embeddings.
% \textbf{ImageNet-1K} \cite{deng2009imagenet} is a large-scale object recognition benchmark with 1,000 classes. We use the standard validation set containing 50,000 images for testing, and the full training set of 1.28 million images for modeling our probabilistic embeddings. \textbf{Flowers-102} \cite{nilsback2008flower102} consists of 102 flower categories, with significant intra-class variation and fine-grained distinctions between classes. We use its official test split with 6,149 images as the test dataset, and 1,020 training images for modeling our probabilistic embeddings. \textbf{Food-101} \cite{bossard2014food101} includes images from 101 food categories. We use the official test split with 25,250 images as test dataset and use the training split of 75,750 images for modeling our probabilistic embeddings.

\begin{figure*}[h]
    \centering
    \includegraphics[width=1.0\textwidth]{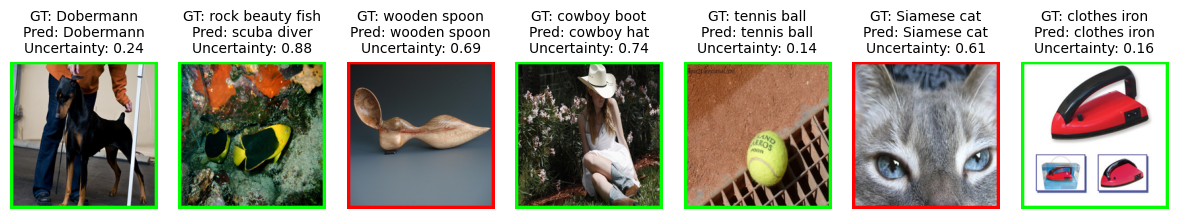} 
    \caption{Visualization of our uncertainty on ImageNet \cite{deng2009imagenet}. 
    Green indicates our method correctly distinguished between correct and error, and red indicates an incorrect distinction. The uncertainty threshold for error rejection is 0.5. 
    % \todo{UPDATE}
    % Green means correct prediction samples, and red means misclassification samples.\TODO{show a random selection of many samples from ImageNet. Make it clearer what green and red mean. Show and discuss what type of samples we do well at, and which ones we seem to fail on.} This can be put in section 4.2 if you need rather than in its own section
    }
    \label{fig:vis}
\end{figure*}

\noindent\textbf{Baselines: }We compare to four standard baselines for uncertainty estimation: maximum cosine similarity, maximum softmax score, entropy of the softmax distribution and temperature scaling~\cite{tu2024empirical}. We additionally compare to 6 state-of-the-art uncertainty estimation methods designed for contrastive VLMs, testing their performance for error detection. This includes methods that model probabilistic embeddings -- ProbVLM~\cite{upadhyay2023probvlm}, PCME++~\cite{chun2024pcmepp}, ProLIP~\cite{chun2025prolip} and BayesVLM~\cite{baumann2024bayesvlm} -- as well as non-probabilistic approaches Zero~\cite{farina2024zero} and TrustVLM~\cite{trustvlm}. Most similar to our approach is TrustVLM~\cite{trustvlm}, which was also designed for the task of error detection and considers intra-class similarity, but does not consider a probability distribution. 
% While the other approaches produce uncertainty estimates, they were not designed for error detection and instead were asse
% test images to training images within the same modality, thereby considering intra-modal similarity. However, it relies solely on cosine similarity without modeling the underlying feature distribution of each class, limiting its ability to distinguish between typical and atypical examples within a class.
% uses different views of one test image. However, it does not model the internal structure or distribution of each class, which lacks the ability to assess whether a test image is truly representative of the predicted class. PCME++ \cite{chun2024pcmepp}, ProLIP \cite{chun2025prolip} and BayesVLM \cite{baumann2024bayesvlm} leverage probabilistic representation and compare the class difference, but these differences are defined across modalities which is between image and text embeddings. In contrast, they overlook intra-modal variation within the image space. TrustVLM \cite{trustvlm} partially addresses this issue by comparing test images to training images within the same modality, thereby considering intra-modal similarity. However, it relies solely on cosine similarity without modeling the underlying feature distribution of each class, limiting its ability to distinguish between typical and atypical examples within a class.

\noindent\textbf{Implementation Details: }
We use the CLIP ViT-B/32 (OpenAI), CLIP ViT-B/16 (OpenAI) and SigLIP (google/siglip-base-patch16-224) as backbones. When applying PCA to feature embeddings, we reduce the feature dimensionality to 128. PCA is fit once on the full set of training embeddings for each dataset and reused across all test images regardless of class. For TrustVLM \cite{trustvlm}, we use their public code and run on ImageNet directly. Due to different training and test data split on Food101 \cite{bossard2014food101}, Flowers102 \cite{nilsback2008flower102}, EuroSAT~\cite{helber2019eurosat} and DTD~\cite{cimpoi14dtd}, we use our data split to keep a fair comparison.

% \TODO{Metrics, datasets, any hyperparameters, baselines, etc.}

\subsection{Uncertainty for Error Detection}
\label{sec:sotaresults}
% \TODO{performance on imagenet plus other datasets compared to baselines}

In Table~\ref{tab:comparison}, we show that our approach achieves state-of-the-art performance for error detection across all 5 datasets. In particular, we observe between a 4-23\% increase in AuROC and a 11-31\% reduction in FPR95 when compared to the next best method. Notably, our approach consistently outperforms the baselines for both the CLIP~\cite{clip} and SigLIP~\cite{siglp} VLMs. These results validate one of the core claims of our paper -- state-of-the-art performance for VLM error detection that is agnostic to the underlying VLM.

Our strong performance can be attributed to two key insights underpinning our method design: (1) capturing intra-class visual consistency, with (2) the use of probabilistic embeddings.
Comparatively, the other baselines that leverage probabilistic embeddings~\cite{chun2024pcmepp, upadhyay2023probvlm, chun2025prolip, baumann2024bayesvlm} perform poorly for the task of error detection, often outperformed by the simple MaxSoftmax baseline -- these approaches were designed for improving top-1 accuracy or confidence calibration tasks, and instead use cross-modal comparison (image-text) rather than our proposed intra-class comparison (image-images). The most competitive baseline is TrustVLM~\cite{trustvlm}. Similar to our approach, TrustVLM utilizes intra-class visual consistency to estimate uncertainty, however it does not leverage probabilistic embeddings and only considers a mean embedding for each class -- our substantial performance gains emphasize the importance of modeling the entire distribution for each class to reliably capture how typical a test image is with respect to the visual feature distribution of its predicted class. 
% Across the prior uncertainty estimation methods from the literature, TrustVLM 
% etween  we observe an This improvement stems from our explicit modeling of class-conditional feature distributions in the image embedding space. Unlike prior approaches that either rely on global similarity (e.g., cosine, softmax) or cross-modal comparison (e.g., image–text distance), our method evaluates how typical a test image is with respect to the internal distribution of its predicted class. This distributional perspective enables more reliable uncertainty estimation. Moreover, our approach can be applied to various vision-language architectures. It achieves SOTA performance on three distinct backbones, demonstrating both effectiveness and robustness across diverse model variants.

% \begin{figure*}[h]
%     \centering
%     \includegraphics[width=1\textwidth]{src/uncertainty_histogram.png} 
%     \caption{The uncertainty histogram for correct and incorrect predictions across six different uncertainty estimation methods on ImageNet-1k \cite{deng2009imagenet}. In each histogram, green bars represent correctly classified samples, while red bars denote misclassified ones.}
%     \label{fig:comp}
% \end{figure*}

% \TODO{clip and siglip?}

% \TODO{change to be about images, histogram has been removed.}
Errors can arise from both epistemic uncertainty (due to out of distribution inputs~\cite{kendall2017uncertainties}) and aleatoric uncertainty (due to ambiguous or noisy inputs~\cite{kendall2017uncertainties}). Figure \ref{fig:vis} shows a random sample of qualitative examples of our uncertainty estimates on ImageNet (additional included in the Supp. Material). Our method primarily appears to capture aleatoric uncertainty, e.g. the noisy data label example of the cowboy boot/hat image and the ambiguous siamese cat image. Our method also shows some evidence of identifying long-tail or out-of-distribution inputs arising from epistemic uncertainty, e.g. the atypical wooden spoon image, which has in fact been classified correctly despite being assigned high uncertainty.

\subsection{Data Dependency of Probabilistic Embeddings}
\label{sec:data-dependency}
% \TODO{How performance changes with \#ims per class}

\begin{figure}[t]
    \centering
    \includegraphics[width=0.5\textwidth]{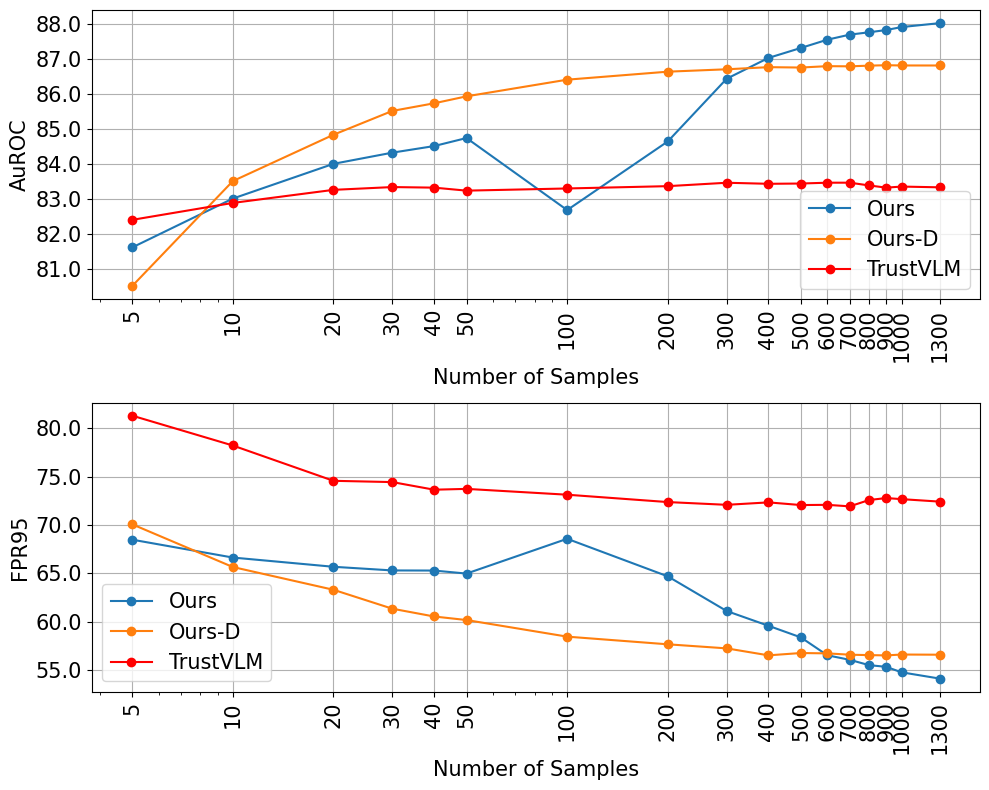} 
    \caption{Testing on ImageNet with a CLIP ViT-B/16, our method achieves SOTA with 10 labeled images per class.
    % The effect of images per class when building sample size on AuROC and FPR95 Performance on the ImageNet dataset. 
    % \todo{UPDATE}
    % \TODO{remove titles, show AuROC and FPR95 only and not AuPR. }
    }
    \label{fig:sample_size_effect}
\end{figure}

% In real-world settings, prior image training sets may vary in size due to data availability. 
To assess the robustness of our method to data availability, we examine how the number of labeled training images per class (used for calculating our dictionary) affects uncertainty estimation performance. As shown in Figure~\ref{fig:sample_size_effect}, both variants of our method achieves SOTA performance with only 10 labeled samples per class. As expected, both variants achieve optimal performance with more data availability (as a greater range of class features can be captured by the class distributions). In the presence of less than 400 labeled images per class, our method using a diagonal covariance matrix for each class Gaussian is preferable to the full covariance Gaussian variant. These findings show that our approach remains competitive even in low-data settings.

\subsection{Importance of Feature Projection for Probabilistic Modeling}
\label{sec:pca}

% \begin{figure}[h]
%     \centering
%     \includegraphics[width=0.5\textwidth]{src/pca-ablation.png} 
%     \caption{Feature projection with PCA is crucial for calculating reliable probabilistic embeddings and our approach achieving state-of-the-art performance.
%     % \todo{UPDATE}
%     }
%     \label{fig:pca-ablation}
% \end{figure}
% \todo{CHANGE THIS BACK TO JUST THE UNCERTAINTY FROM THE INTRA-CLASS MODELLING}

In Table~\ref{tab:pca-ablation}, we evaluate the effect of feature projection in our method by comparing our intra-class uncertainty with and without the PCA feature projection (note that here we do not include the inter-modal softmax). As introduced in Sec.~\ref{sec:method-pca}, PCA is applied to reduce the dimensionality of CLIP image features and mitigate ill-conditioned covariance matrices. After incorporating PCA, performance on ImageNet, Food101 and EuroSAT in particular are improved by 2-8\% AuROC and 15-24\% FPR95. 

\begin{table}[t]
\centering
\setlength{\tabcolsep}{1.25pt}
{\fontsize{8}{10}\selectfont 
\caption{Error detection performance of our method with (w) and without (w/o) PCA.}
\label{tab:pca-ablation}
\begin{tabular}{l|cc|cc|cc|cc|cc}
\hline
\multirow{2}{*}{Metric} & \multicolumn{2}{c|}{ImageNet} & \multicolumn{2}{c|}{Flowers102} & \multicolumn{2}{c|}{Food101} & \multicolumn{2}{c|}{EuroSAT} & \multicolumn{2}{c}{DTD} \\
 & w/o & w & w/o & w & w/o & w & w/o & w & w/o & w \\
\hline
% AuROC & 86.18 & 87.52 & 97.84 & 97.59 & 90.38 & 92.58 & 94.94 & 96.37 & 92.18 & 91.13 \\
% FPR95 & 60.08 & 54.26 & 3.15  & 4.86  & 48.43 & 33.98 & 32.29 & 16.39 & 38.06 & 46.43 \\
AuROC & 82.46 & 87.71 & 95.62 & 95.39 & 84.78 & 93.18 & 95.69 & 98.44 & 88.61 & 86.95 \\
FPR95 & 74.70 & 54.98 & 3.19  & 3.64  & 59.80 & 35.58 & 20.17 & 4.26 & 100.00 & 100.00\\
\hline
\end{tabular}}
\end{table}

\begin{figure}[h]
    \centering
    \includegraphics[width=0.5\textwidth]{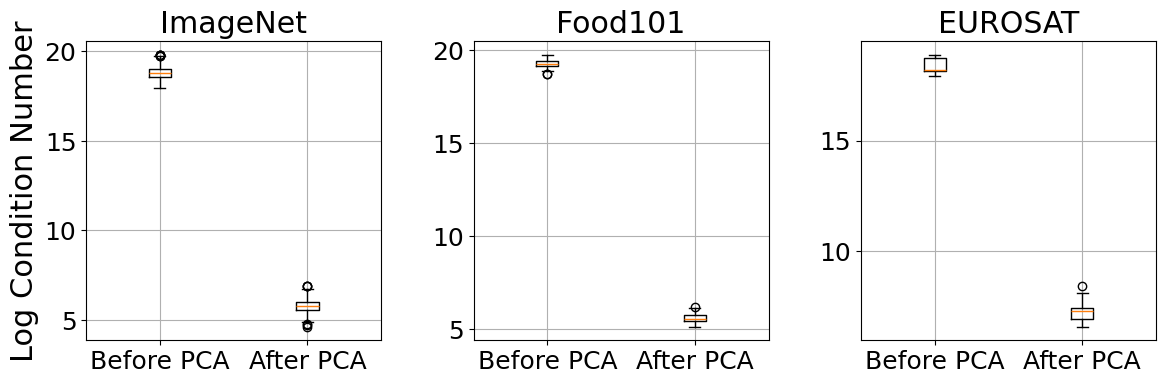} 
    \caption{For our probabilistic embeddings,  application of PCA mitigates ill-conditioned covariance matrices, measured by the log condition number of the covariance matrices for each class. 
    % \todo{UPDATE}
    % TODO{show for flowers102 and food101 as well. Remove title. increase font size so that it is readable in the figure.}
    % \todo{REMOVE Flowers102 and DTD and put them in the Supp Material.}
    }
    \label{fig:condition}
\end{figure}

In addition, in Figure~\ref{fig:condition}, we show that the use of PCA on the feature embeddings substantially reduces the condition number of the covariance matrices for each class probabilistic embedding in ImageNet, Food101 and EUROSAT (see Supp. Material for Flowers102 and DTD). Higher condition numbers indicate ill-conditioned covariance matrices that are prone to numerical instability. This result, combined with the substantial performance increase on these datasets when applying PCA, supports one of the core claims of our paper -- PCA-based feature projection is crucial for allowing each class distribution to be modeled effectively as it mitigates ill-conditioned covariance matrices in the VLM feature space.

\subsection{Robustness to Distribution Shift}
\label{sec:dist-shift}

We explore the robustness of our method under two types of data distribution shift \cite{zhang2023dive}: feature shift (also referred to as covariate shift) and class label shift.

\noindent\textbf{Feature shift:} 
% In this experiment, we test our performance when there is some feature distribution shift between the images used to create our dictionary and the images seen at test time. 
We create our dictionary using the ImageNet training data, and then evaluate performance on three ImageNet variants with different levels of feature shift: ImageNetV2~\cite{recht2019imagenetv2} (image source distribution shift through newly collected real images), ImageNet-C~\cite{hendrycks2018imagenetc} (synthetic distribution shift via image corruptions)
, and ImageNet-A~\cite{hendrycks2021nae} (adversarial natural distribution shift via naturally occurring hard examples). 
% \todo{ADD IN TABLE AND EXPLAIN RESULTS.}

As shown in Table~\ref{tab:distribution-shift}, our method achieves the best AuROC and FPR95 performance on both ImageNetV2 and ImageNet-C, demonstrating robustness to feature shift introduced by image sources and synthetic image corruptions. However, ImageNet-A poses a significant challenge for both our approach and TrustVLM~\cite{trustvlm}, where error detection on natural adversarial images is tested -- in this case, the simple baseline of MaxSoftmax achieves best performance. As noted in~\cite{hendrycks2021nae}, the image distribution shift between ImageNet and ImageNet-A is significant, highlighting the limits of error detection methods that rely on data for uncertainty estimation.

\begin{table}[t]
\centering
\setlength{\tabcolsep}{1pt}
{\fontsize{7}{10}\selectfont 
\caption{Error detection performance under feature shift, where training data is from ImageNet and methods are evaluated on ImageNet variants with CLIP ViT-16.
% Results of feature shift. Zero~\cite{farina2024zero}, ProLIP~\cite{chun2025prolip} and TrustVLM*~\cite{trustvlm} are tested on CLIP ViT-B/16. Others are tested on CLIP ViT-B/32. 
% \todo{THESE NEED TO ALL BE THE SAME BACKBONE WHICH SHOULD BE SPECIFIED IN THIS CAPTION.}
}
\label{tab:distribution-shift}
\begin{tabular}{c|ccc|ccc|ccc}
\hline
\multirow{3}{*}{\textbf{Method}}   & \multicolumn{3}{c|}{\textbf{ImageNet-V2}} & \multicolumn{3}{c|}{\textbf{ImageNet-C}} & \multicolumn{3}{c}{\textbf{ImageNet-A}} \\
 & AuROC  & AuPR & FPR95 & AuROC & AuPR & FPR95 &   AuROC & AuPR & FPR95 \\
&  ($\uparrow$)  & ($\uparrow$) & ($\downarrow$)  & ($\uparrow$)  & ($\uparrow$) & ($\downarrow$) & ($\uparrow$)  & ($\uparrow$) & ($\downarrow$) \\
\hline
% MaxCosine  & 67.46 & 76.06 & 85.47  & 67.98 & 58.38 & 86.26 & 59.65 & 36.04 & 91.00  \\
% MaxSoftmax & 80.87 & 87.36 & 71.33 & 81.31 & 75.47 & 68.45  & \textbf{71.90} & \textbf{51.90} & \textbf{81.05} \\
% Entropy    & 78.39 & 85.96 & 77.12  & 78.96 & 73.05 & 72.87 & 70.62 & 50.52 & 81.22  \\
MaxCosine  & 67.84 & 79.59 & 83.55 & 65.99 & 59.88 & 87.41 & 64.29 & 60.95 & 86.67  \\
MaxSoftmax & 81.30 & 89.83 & 71.00 & 81.48 & 78.34 & 68.32 & \textbf{76.14} & 74.63 & 77.74  \\
Entropy    & 78.78 & 88.62 & 78.08 & 78.84 & 75.72 & 72.86 & 74.82 & 73.40 & \textbf{77.66} \\
% TempScaling~\cite{tu2024empirical} & 81.01 & 87.48 & 71.16  & 81.24 & 75.39 & 68.54 & 71.87 & 51.81 & 81.16 \\
TempScaling~\cite{tu2024empirical} & 81.53 & 89.95 & 70.43 & 81.65 & 78.57 & 68.14 & 76.12 & \textbf{74.84} & 78.20 \\
% ProbVLM~\cite{upadhyay2023probvlm} & 53.01 & 63.20 & 93.97 \\
% PCME++~\cite{chun2024pcmepp} & 79.01 & \textbf{90.70} & 71.89  \\
% Zero~\cite{farina2024zero} & 78.82 & 90.21 & 74.80 3 \\
% ProLIP~\cite{chun2025prolip} & 73.95 & 87.43 & 77.13 \\
% BayesVLM~\cite{baumann2024bayesvlm} & 80.53 & 87.12 & 71.99  \\
TrustVLM \cite{trustvlm} & 83.03 & 89.68 & 78.76 & 82.12 & \textbf{83.73} & 71.99 & 64.28 & 49.53 & 92.92 \\
 % \rowcolor{gray!25}Ours-D & 84.22 & 89.53 & 64.51 & \textbf{82.84} & 77.03 & \textbf{65.30} & 64.51 & 42.19 & 83.11 \\
 % \rowcolor{gray!25}Ours   & \textbf{85.37} & \textbf{90.34} & \textbf{63.16} & 82.41 & 75.95 & 65.72 & 63.51 & 39.49 & 81.56 \\
  \rowcolor{gray!25}Ours-D & 85.32 & 91.91 & 62.27 & \textbf{83.51} & 80.12 & \textbf{64.09} & 70.34 & 67.53 & 77.99   \\
 \rowcolor{gray!25}Ours    & \textbf{86.19} & \textbf{92.42} & \textbf{59.00} & 82.45 & 78.41 & 65.06 & 68.83 & 64.74 & 78.38   \\
\hline
\end{tabular}
}
\end{table}

\noindent\textbf{Class label shift:} We test our performance when the VLM query labels do not exactly match the keys in our class probabilistic embedding dictionary, simulating a coarse-to-fine mismatch setting on the ImageNet-1k dataset \cite{deng2009imagenet}. We assume that every query class is represented in the dictionary, but that the text input may not exactly match, utilizing
% The result shown in Table \ref{tab:lv1} and Table \ref{tab:lv2} supports a core claim of our paper that modeling intra-class feature distributions enables more robust uncertainty estimation, even under conditions of semantic mismatch between retrieval and test label spaces.
% Models often encounter a label semantic shift, where the granularity of labels changes between retrieval and test dataset. To evaluate the robustness of the uncertainty estimation under such label shift, we simulate a coarse-to-fine mismatch setting on the ImageNet-1k dataset \cite{deng2009imagenet}. 
% We utilize 
the WordNet hierarchy \cite{fellbaum1998wordnet} to group the 1,000 fine-grained ImageNet labels into coarser categories. We construct two levels: the 1st-level superclass containing 548 labels and the 2nd-level (grand) superclass containing 299 labels. The test set labels are replaced with their corresponding superclasses, while the dictionary's class labels remain the original fine-grained labels. To adapt our method to this setting, we retrieve the labels from the dictionary which are among the top-K most semantically similar classes with the test class, where similarity is measured by cosine similarity in the CLIP text embedding space. The images with these labels are then used to construct probabilistic embeddings for the corresponding superclass. The number K is selected based on the ratio between the number of classes in the dictionary and the number of test-time labels, calculated by
\begin{equation}
\begin{aligned}
    K = \left\lfloor \frac{N_{\text{retrieval}}}{N_{\text{test}}} \right\rceil.
\end{aligned}
\label{eq:selec-k}
\end{equation}

% Table~\ref{tab:lv1} and Table~\ref{tab:lv2} 
Table~\ref{tab:lv1lv2} shows our method outperforms all baselines in all metrics under both label shift levels. This suggests that our class-specific probabilistic embeddings capture uncertainty more reliably than similarity-based baselines. Compared to TrustVLM-LS, which also attempts to estimate uncertainty under label shift using mean embeddings of top-K classes, our method demonstrates improved robustness by modeling the full feature distribution rather than only relying on mean feature. 
% In addition, TrustVLM suffers accuracy degradation under label shift. However, our accuracy does not go lower than the standard VLM because we do not change prediction mechanism and are fully posthoc uncertainty method. 
This result supports the final core claim of our paper that our method exhibits robustness to label shift, and can continue to exhibit SOTA performance when the probabilistic embedding labels have some discrepancy with the test query labels.
% This highlights the advantage of distribution-aware uncertainty modeling in handling semantic mismatches between dictionary and test label spaces.

\begin{table}[t]
\centering
\caption{Results of label shift on the 1st and 2nd level superclasses. 
For 1st-level, we set $K=2$. For 2nd-level, we set $K=3$.}
\setlength{\tabcolsep}{2pt}
{\fontsize{7}{10}\selectfont 
\label{tab:lv1lv2}
\begin{tabular}{c|cccc|cccc}
\hline
\multirow{2}{*}{Method} & \multicolumn{4}{c|}{1st level} & \multicolumn{4}{c}{2nd level} \\
\cline{2-9}
& AuROC & AuPR & FPR95 & Acc & AuROC & AuPR & FPR95 & Acc \\
\hline
MaxCosine    & 62.58 & 43.87 & 87.72 & \textbf{32.53} & 63.53 & 27.90 & 87.06 & \textbf{19.15} \\
MaxSoftmax   & 71.78 & 55.46 & 81.05 & \textbf{32.53} & 67.89 & 31.86 & 82.73 & \textbf{19.15} \\
Entropy      & 69.78 & 53.69 & 83.76 & \textbf{32.53} & 67.43 & 31.51 & 83.55 & \textbf{19.15} \\
% \hline
TempScaling~\cite{tu2024empirical} & 71.76 & 55.56 & 81.09 & 32.51 & 67.58 & 31.68 & 82.81 & 19.14 \\
TrustVLM-LS~\cite{trustvlm} & 74.00 & 57.42 & 87.40 & 26.72 & 68.06 & 28.60 & 91.99 & 13.81 \\
\rowcolor{gray!25}Ours-D-LS & 77.15 & 62.39 & 75.44 & \textbf{32.53} & 70.47 & 34.05 & \textbf{79.86} & \textbf{19.15} \\
\rowcolor{gray!25}Ours-LS & \textbf{77.34} & \textbf{62.69} & \textbf{75.18} & \textbf{32.53} & \textbf{70.64} & \textbf{34.28} & 80.83 & \textbf{19.15} \\
\hline
\end{tabular}}
\end{table}

\subsection{Sensitivity to Uncertainty Threshold}
\label{sec:sensitive}
\begin{figure}[h]
    \centering
    \includegraphics[width=0.5\textwidth]{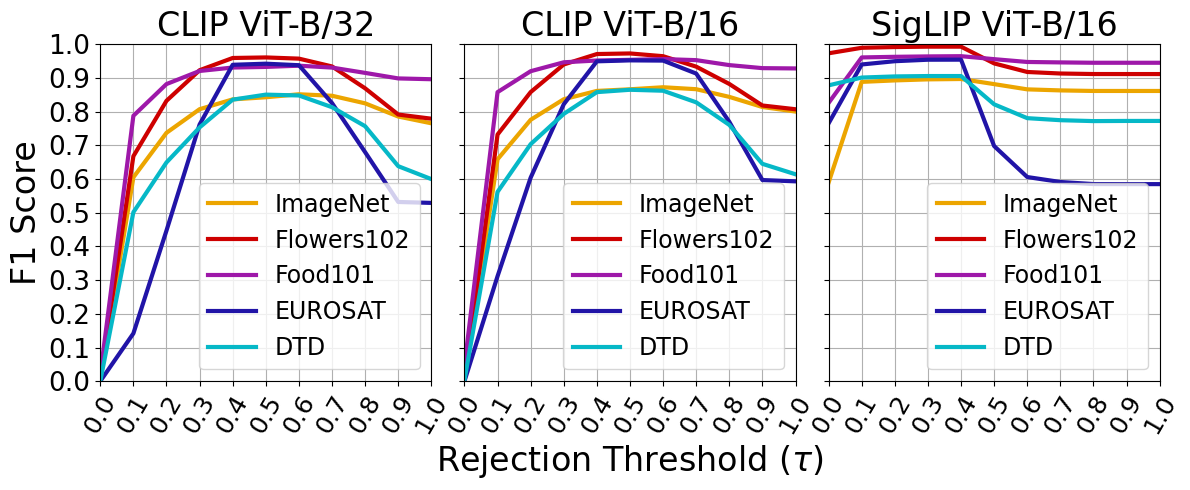} 
        \caption{Sensitivity of our approach for error detection across a range of uncertainty rejection thresholds.
    }
    \label{fig:sensitive-analysis}
\end{figure}

To further examine the robustness of our approach, we explore the sensitivity of error detection performance to the rejection threshold $\tau$, which determines when a prediction is flagged as an expected error. As shown in Figure~\ref{fig:sensitive-analysis}, across all five datasets, the F1 score of our method remains at a stable peak between 0.4 and 0.6 for CLIP backbones and between 0.1 and 0.4 for the SigLIP backbone. This indicates that our method does not require precise tuning of $\tau$, and a value of 0.4 can be used for both backbones to consistently achieve strong performance for error detection. 

\color{black}

\section{Conclusion}
% \TODO{read and check consistent terminology with entire paper, check matching the claims we make in introduction}
In this paper, we present a training-free, post-hoc method for uncertainty estimation in vision-language models, focusing on error detection. Our method estimates uncertainty by combining inter-modal uncertainty with a measure of intra-class feature consistency captured by distributions in the visual embedding space.
Although the proposed method achieves state-of-the-art performance across multiple benchmarks, its reliance on data for building the class-wise probabilistic distributions introduces two challenges that should be explored in future work: (1) performance degrades under feature shift on adversarially curated natural examples, indicating susceptibility in boundary-adjacent, hard regions of the input space; and (2) this method may become intractable when the distributions must be built on compute-constrained edge devices.
% . Future work will incorporate epistemic-aware extensions, such as Bayesian or ensemble-based distribution estimation, and strengthen adversarial robustness via robust optimization and adversarial data augmentation.
% \section*{Acknowledgment}
% The authors acknowledge continued support from the Queensland University of Technology (QUT) through the Centre for Robotics. 
{
    \small
    \bibliographystyle{ieeenat_fullname}
    \bibliography{main}
}

\end{document}

\maketitle
\appendix

We provide additional results and analyses to complement the findings in the main paper. We first report the classification accuracy of different methods under various backbones. We then present supplementary results on PCA-based feature projection to further demonstrate its effect on alleviating ill-conditioned covariance matrices. Finally, we show additional qualitative visualizations of our uncertainty estimates on ImageNet to illustrate the effectiveness of our approach.

\section{Accuracy of different methods}
Table~\ref{tab:acc-vit32}, \ref{tab:acc-vit16} and \ref{tab:acc-siglip} show the accuracy of different methods under different backbones.
\begin{table}[h]
\setlength{\tabcolsep}{3pt}
{\fontsize{8}{10}\selectfont 
    \centering
    \caption{Accuracy of different methods (CLIP ViT-B/32)}
    \label{tab:acc-vit32}
    \begin{tabular}{c|ccccc}
      \hline
      Method  & ImageNet  & Flowers102 & Food101 & EuroSAT & DTD \\
      \hline
      MaxCosine   & 62.02 & 63.83 & 81.22 & 35.99 & 42.85 \\
      MaxSoftmax  & 62.02 & 63.83 & 81.22 & 35.99 & 42.85 \\
      Entropy   & 62.02 & 63.83 & 81.22 & 35.99 & 42.85 \\
      TempScaling~\cite{tu2024empirical}   & 62.02 & 63.83 & 81.22 & 35.99 & 42.85 \\
      ProbVLM~\cite{upadhyay2023probvlm}   & 61.78 & 59.29 & 81.25 & 35.88 & 41.84 \\
      PCME++~\cite{chun2024pcmepp} & 77.61 & 82.05 & 89.05 & 98.75 & 75.65 \\
      BayesVLM~\cite{baumann2024bayesvlm} & 61.30 & 64.90 & 80.61 & 30.68 & 30.44 \\
      Ours  & 62.02 & 63.83 & 81.22 & 35.99 & 42.85 \\
      \hline
    \end{tabular}}
\end{table}

\begin{table}[h]
\setlength{\tabcolsep}{3pt}
{\fontsize{8}{10}\selectfont 
    \centering
    \caption{Accuracy of different methods (CLIP ViT-B/16)}
    \label{tab:acc-vit16}
    \begin{tabular}{c|ccccc}
      \hline
      Method  & ImageNet  & Flowers102 & Food101 & EuroSAT & DTD \\
      \hline
      MaxCosine   &  66.73& 67.69&   88.65& 42.17&44.33\\
      MaxSoftmax   &   66.73& 67.69&  88.65& 42.17&44.33\\
      Entropy   &   66.73& 67.69&  88.65& 42.17&44.33\\
      TempScaling~\cite{tu2024empirical}   & 66.74& 67.64&   88.65& 42.17&44.33\\
      Zero~\cite{farina2024zero}   & 71.19  & 68.24 & 88.36 & 42.30 & 45.80 \\
      ProLIP++~\cite{chun2025prolip} & 73.70 & 78.84 & 90.94 & 44.93 & 63.36 \\
      TrustVLM~\cite{trustvlm} & 65.04 & 99.01 & 86.88 & 81.06 & 72.28 \\
      Ours  &  66.73& 67.69&  88.65& 42.17&44.33\\
      \hline
    \end{tabular}}
\end{table}

\begin{table}[h]
\setlength{\tabcolsep}{3pt}
{\fontsize{8}{10}\selectfont 
    \centering
    \caption{Accuracy of different methods (SigLIP ViT-B/16)}
    \label{tab:acc-siglip}
    \begin{tabular}{c|ccccc}
    \hline
      Method  & ImageNet  & Flowers102 & Food101 & EuroSAT & DTD \\
      \hline
      MaxCosine   & 75.67 & 83.77 & 89.63 & 41.35 & 62.88\\
      MaxSoftmax   & 75.67 & 83.77 & 89.63 & 41.35 & 62.88\\
      Entropy   & 75.67 & 83.77 & 89.63 & 41.35 & 62.88\\
      TempScaling~\cite{tu2024empirical}   & 75.67 & 83.77 & 89.63 & 41.33 & 62.94\\
      Zero~\cite{farina2024zero}   & 77.71  & 82.35 & 88.73 & 29.00 & 64.01 \\
      BayesVLM~\cite{baumann2024bayesvlm} & 75.68 & 81.77 & 89.58 & 41.30  & 48.58 \\
      TrustVLM~\cite{trustvlm} & 68.68 & 99.07 & 88.20 & 80.62 & 71.99 \\
      Ours & 75.67 & 83.77 & 89.63 & 41.35 & 62.88 \\
      \hline
    \end{tabular}}
\end{table}

\section{Additional Results on PCA-based Feature Projection}
\begin{figure}[h]
    \centering
    \includegraphics[width=0.5\textwidth]{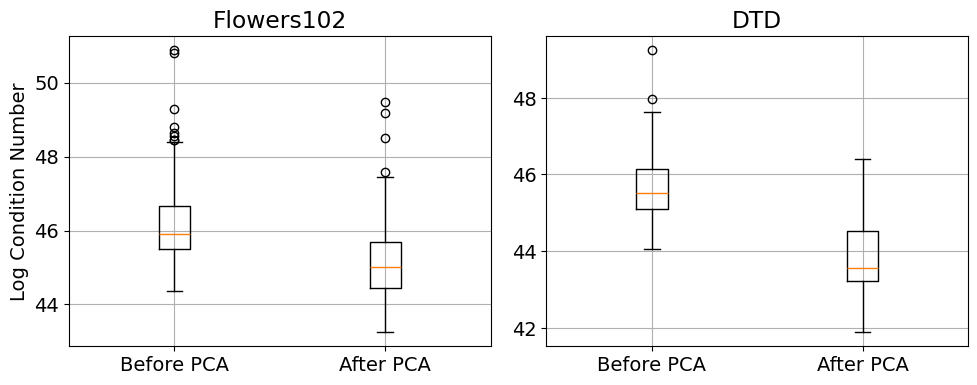} 
    \caption{
    Log condition numbers of class-wise covariance matrices before and after PCA on Flowers102 and DTD. PCA reduces the condition number, indicating our probabilstic embeddings effectively alleviate the ill-conditioned covariance matrices.
    % For our probabilistic embeddings,  application of PCA mitigates ill-conditioned covariance matrices, as measured by a boxplot of the log condition number of the covariance matrices for each class. 
    }
    \label{fig:condition-2}
\end{figure}

To complement the results presented in Sec.~4.3, we further report the log condition numbers of class-wise covariance matrices for Flowers102~\cite{nilsback2008flower102} and DTD~\cite{cimpoi14dtd} datasets. As shown in Fig.~\ref{fig:condition-2}, applying PCA consistently reduces the condition number, indicating that feature projection mitigates ill-conditioning across datasets of different scales and domains.

\begin{figure*}[h]
    \centering
    \includegraphics[width=1.0\textwidth]{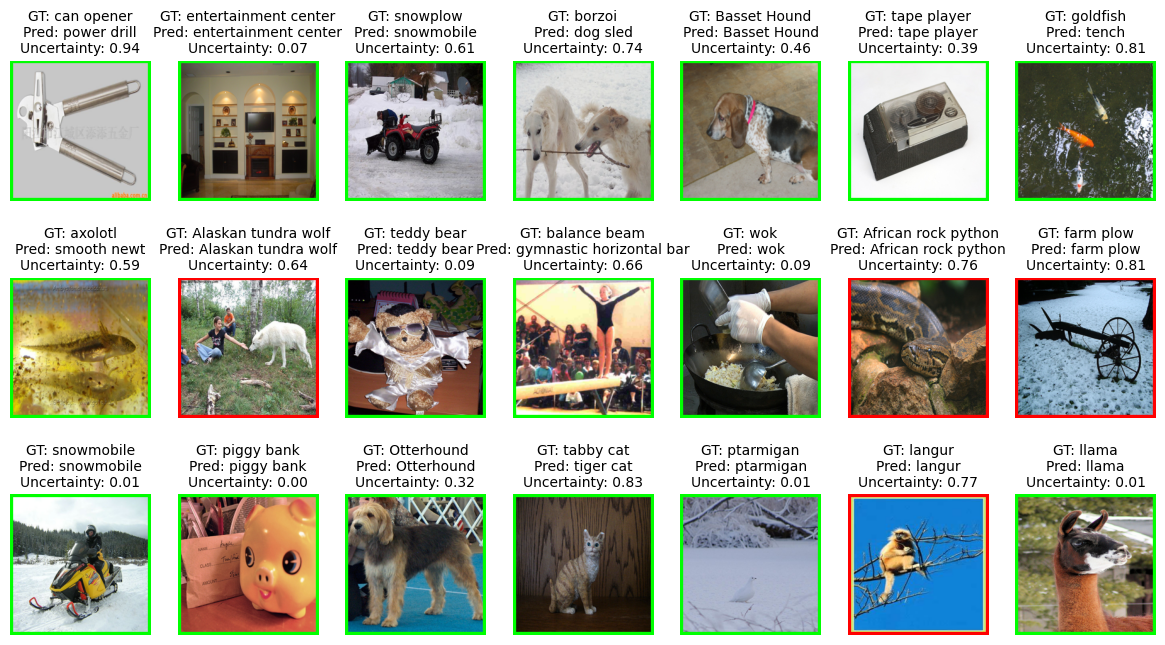} 
    \caption{Visualization of our uncertainty on ImageNet \cite{deng2009imagenet}. 
    Green means examples where are uncertainty has correctly distinguished between correct and error, and red means when our uncertainty failed to distinguish them. The threshold of uncertainty is 0.5.
    }
    \label{fig:vis-appendix}
\end{figure*}

\section{Additional Quantitative Examples}
Figure~\ref{fig:vis-appendix} shows additional quantitative examples on ImageNet~\cite{deng2009imagenet}.

% \input{sec/rebuttal-tables/reb-comp-scaling}

%%%%%%%%% BODY TEXT - ENTER YOUR RESPONSE BELOW

%%%%%%%%% REFERENCES
{
    \small
    \bibliographystyle{ieeenat_fullname}
    \bibliography{main}
}